# Segmenting Brain Tumors with Symmetry

**Hejia Zhang**[*]
Princeton University

**Xia Zhu**
Intel Labs

**Theodore L. Willke**
Intel Labs

## Abstract

We explore encoding brain symmetry into a neural network for a brain tumor segmentation task. A healthy human brain is symmetric at a high level of abstraction, and the high-level asymmetric parts are more likely to be tumor regions. Paying more attention to asymmetries has the potential to boost the performance in brain tumor segmentation. We propose a method to encode brain symmetry into existing neural networks and apply the method to a state-of-the-art neural network for medical imaging segmentation. We evaluate our symmetry-encoded network on the dataset from a brain tumor segmentation challenge and verify that the new model extracts information in the training images more efficiently than the original model.

## 1 Introduction

Brain tumor segmentation using Magnetic Resonance Imaging (MRI) plays an important role in the diagnosis and treatment of brain tumors. Traditionally, this is done manually. However, the manual segmentation is time-consuming, prone to human error, and requires significant expertise [1]. Thus, there is an increasing interest in developing automatic brain tumor segmentation methods. Some methods [2, 3] extract features with medical domain knowledge and use traditional classifiers such as random forest and support vector machines. Recently, deep learning methods [4, 5, 6, 7, 8, 9] show promising results in this task and offer better performance in general. We observe that many of these deep learning methods can be used in other biomedical image segmentation tasks or even semantic segmentations as well. This flexibility can be an advantage but, on the other hand, it implies that these methods make little to no use of the specific properties of the brain or brain tumors. Specific properties could include the structure of normal brain tissue, the texture of brain tumors, and so on. Encoding these properties into the deep learning methods has the potential to extract features that are more relevant to the brain tumor segmentation task.

Here we focus on encoding structural property about the brain. One of the universal characteristics of brain structure is the bilateral symmetry of the brain. More precisely, the left hemisphere and right hemisphere of a healthy brain are symmetric at a high level of abstraction, and the presence of any high-level asymmetry usually implies abnormality. Thus, in the brain tumor segmentation task, we would like to pay more attention to asymmetric portions, which are more likely to be tumor regions.

In this work, we propose a method to encode the brain symmetry into neural networks for brain tumor segmentation. We apply this method to encode brain symmetry into one of the state-of-the-art neural networks for medical imaging segmentation. The new model and the original model is compared on a popular brain tumor segmentation challenge dataset.

## 2 Symmetry Encoded Neural Network

Now we explore how to encode symmetry into neural networks. The basic idea is to compute the difference between the original brain image and the flipped brain image at a high level and concentrate on the difference when extracting features. In a neural network, we implement this idea by subtracting the original and flipped image at some feature map layer. The difference is concatenated with the feature maps of the original image at that level to preserve the information of the original image as well. We can apply this idea to encode symmetry in existing neural networks for this task. However, as the brain symmetry is at a high level of abstraction, we would like to perform the subtraction after several layers of convolution and pooling. Thus, this idea is most appropriate for neural networks with consecutive convolution and pooling layers to extract features at different abstraction levels.

---

[*]This work was done while Hejia Zhang was an intern at Intel Labs

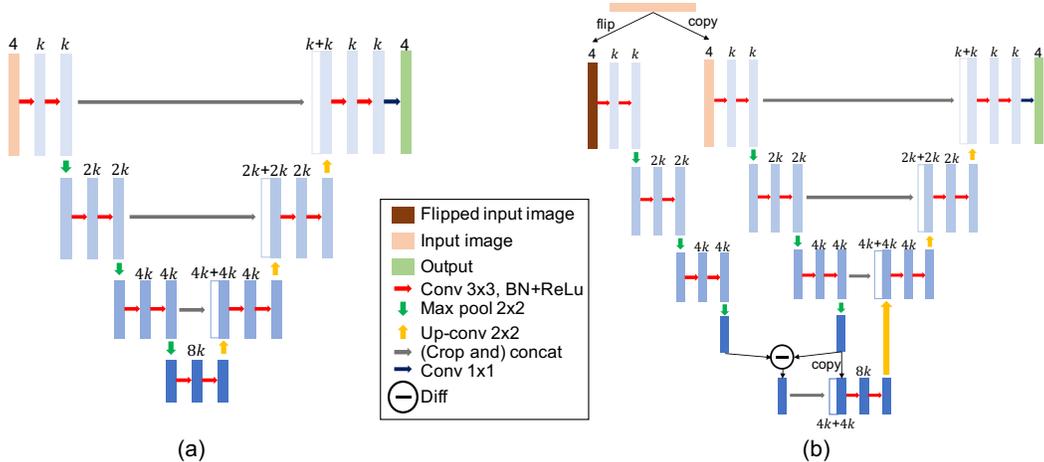

Figure 1: **(a)** U-net used as the baseline. **(b)** Symmetry-encoded U-net. The flipped image branch and the original image branch share the same filter weights. $k$ is the number of feature maps.

| k | 4 | 8 | 16 | 32 |
|---|---|---|---|---|
| **Num. of Param. in U-Net** | 30904 | 121820 | 483940 | 1929332 |
| **Num. of Param. in Symmetry-encoded U-Net** | 35636 | 140488 | 558128 | 2225152 |
| **Ratio (Symm/U-Net)** | 1.15 | 1.15 | 1.15 | 1.15 |

Table 1: Number of parameters in the standard U-Net and the Symmetry-encoded U-Net.

Here we use U-Net [4] as an example. It is one of the state-of-the-art neural networks for medical imaging segmentation and has convolution and pooling layers for feature extraction. The architecture of a standard U-Net is shown in Fig. 1(a). U-Net is a fully convolutional network that can be trained end-to-end. The downward path consists of a series of convolution and pooling layers that extract features of different resolutions. The upward path contains convolution and up-convolution layers that increase the resolution of outputs. The skip shortcuts transfer information from the downward path to the upward path to help define more precise boundaries in the segmentation. The number of feature maps is doubled at each layer in the downward path and halved at each layer in the upward path as in [4]. Both the input and output have $4$ channels (details discussed in § 3). We encode symmetry into U-Net by using two branches that share the same filter weights in the downward path. The two branches take in the original and the flipped brain image, respectively. The difference between feature maps from these two branches is computed after several layers and concatenated with the feature maps of the original image. This symmetry-encoded U-Net is illustrated in Fig. 1(b). Note that the filter weights of two branches are shared, so the number of parameters in this network is only $1.15X$ of the number of parameters in the standard U-Net (shown in Tab. 1). Also, note that our model is not the same as data augmentation with flipped images. Our model sees the original and flipped images at the same time and explicitly utilizes their differences in high-level features maps, whereas the data augmentation method sees the original and flipped images, respectively, and treats them as independent training images.

## 3  Experiments

We use the *Multimodal Brain Tumor Segmentation Challenge* (BraTS) [10] to determine if our symmetry-encoded model (Fig. 1(b)) has an improved performance compared to the standard U-Net (Fig. 1(a)) for the brain tumor segmentation task.

*Dataset:*  We use the training set of the BraTS2017 dataset [10]. It contains multimodal MRI scans of 285 subjects with brain tumors from clinics. The four modes are T1, post-contrast T1 (T1Gd), T2, and T2 fluid attenuated inversion recovery (FLAIR) [11]. Brain image from each patient of each mode is a 3D matrix of size $240 \times 240 \times 155$. All the images are anatomically registered to the same brain template. Images are segmented manually by experts into four classes: the background or healthy part (class 0), the Gd-enhancing tumor (ET, class 1), the peritumoral edema (ED, class 2), and the necrotic and non-enhancing tumor (NCR/NET, class 3). The 285 subjects are split into two groups based on the severity of brain tumors. The more severe group called high-grade gliomas (HGG) [12] has 210 subjects, and the other called lower grade gliomas (LGG) has 75 subjects. We use the HGG group in this paper.

*Preprocessing:*  a) As our network takes 2D images, we convert each 3D brain image as 155 2D slices of size $240 \times 240$. b) We observe that on average, 99.1% of the pixels are annotated as class 0 (healthy tissue), so we throw away any image with class 0 only. We also crop the image size to $156 \times 192$ to focus on brain itself. To ensure the flipped image is aligned with the original image, the cropped image has its axis of symmetry at approximately the middle of the image. c) Each image is scaled to $[0, 1]$ in intensity and matched to a template using histogram matching [13]. The template is the average image across all images. After the preprocessing steps, each image is of size $156 \times 192 \times 4$ where we put each mode in a different channel.

*Experiment Details:*  We randomly partition the 210 subjects into two parts, with 80% for training and 20% for validation. The experiment is repeated 5 times with different partitioning and the averaged result is reported. We use the weighted cross-entropy loss for training. As 95.1% of the pixels are still class 0 on average even after preprocessing, the classes are still highly imbalanced. Thus, we apply median frequency balancing [14] to weight the cross-entropy loss. The idea is to penalize more when a small class gets errors. The loss is defined as

$$-\sum_{c=1}^{C} \alpha_c \sum_{p \in c} \sum_{i=1}^{C} t_{pi} log(y_{pi})$$

, where $C$ is number of classes, $p$ is pixel, $t_{pi}$ and $y_{pi}$ are the true binary label and the output from softmax of class $i$ for pixel $p$, and $\alpha_c$ is the class weight of class $c$ defined as $\alpha_c = \frac{\text{median freq across all classes}}{\text{freq of class } c}$. The learning algorithm is Adam with Nesterov momentum (Nadam) [15] with learning rate $0.001$. The evaluation metric is the average Dice score [16, 17] across 4 classes. The Dice score is good at evaluating the quality of a segmentation result when the classes are imbalanced. For each class, the Dice score is defined as

$$\frac{2|T \cap P|}{|T| + |P|}$$

, where T and P are true and predicted binary labels, and $|\cdot|$ is the area.

We tried $k = 4, 8, 16, 32$, where $k$ is number of feature maps of the first layer of neural network as shown in Fig. 1. We evaluate Dice scores on validation data whenever 8000 training images are processed. When the average Dice score across different tumor classes does not increase for five consecutive evaluations on validation data, an experiment is marked as converged. The number of epochs it took to converge and the average Dice scores of different tumor classes with different $k$ on the validation set are shown in Fig. 2. An example segmentation output from the symmetry-encoded U-Net is shown in Fig. 3.

We see that the symmetry-encoded U-Net consistently takes fewer epochs to converge. This suggests that the symmetry-encoded model can more efficiently extract information from the training images and eventually train the model with fewer epochs while achieving similar segmentation performance. When k is 4 or 8, the new model also gets better Dice scores compared with the U-Net. When k is 16 or 32, the two models have similar Dice scores. A possible reason is that when k is small, the new model extracts more relevant features than the baseline U-Net, but as k gets larger, the U-Net starts to get the redundancy to include both relevant features and other features.

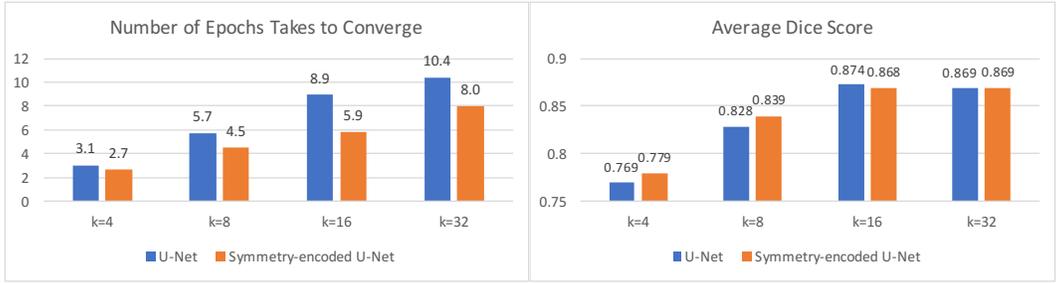

Figure 2: Results from U-Net and Symmetry-encoded U-Net on BraTS2017 dataset. The experiment is repeated 5 times based on different partitions of training and validation sets. The averaged validation results are reported here. **Left:** Number of epochs takes to converge. The criterion for convergence is described in the text. **Right:** Average Dice score across 4 classes on validation data. The Dice score is computed right after convergence.

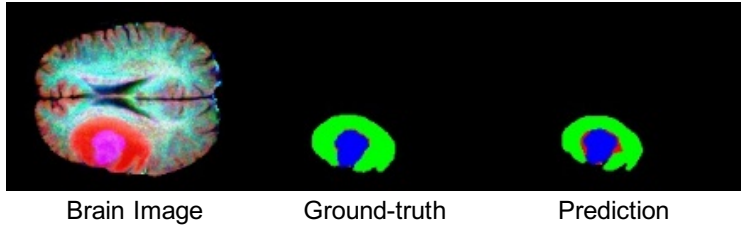

Figure 3: An example of segmentation output image from the symmetry-encoded U-Net. Red: class 1; Green: class 2; Blue: class 3.

## 4 Discussion and Conclusion

We proposed a method to encode brain symmetry information into neural networks for brain tumor segmentation. We encode symmetry into U-Net by adding a branch with flipped brain images to the downward path and subtracting between two branches at a high level of abstraction. This technique can also be extended to other neural networks with successive convolution and pooling layers. We compared the performance of the symmetry-encoded U-Net and the standard U-Net on a popular brain tumor segmentation dataset. The new model takes fewer epochs to converge and shows a higher Dice score when a small number of feature maps are used. The result suggests that encoding symmetry into the neural network helps extract features that are more relevant to the brain tumor segmentation task. In essence, our method forces the network to pay more attention to high-level asymmetric portions, which are more likely to be tumor regions. The relationship between asymmetric portions and tumor regions may be learned implicitly with the standard U-Net, but will take more effort. Our method encodes this information and spares the effort.

The idea can be easily extended to 3D U-Net and also can be extended to encode task-specific medical domain knowledge into neural networks for medical applications.